\documentclass[10pt,twocolumn,letterpaper]{article}

\usepackage[pagenumbers]{cvpr} 

\usepackage{graphicx}
\usepackage{amsmath}
\usepackage{amssymb}
\usepackage{booktabs}
\usepackage{soul}
\usepackage[table]{xcolor}
\usepackage{balance}
\usepackage{multirow}

\usepackage[pagebackref,breaklinks,colorlinks]{hyperref}

\usepackage[capitalize]{cleveref}
\crefname{section}{Sec.}{Secs.}
\Crefname{section}{Section}{Sections}
\Crefname{table}{Table}{Tables}
\crefname{table}{Tab.}{Tabs.}

\def\cvprPaperID{13133} 
\def\confName{CVPR}
\def\confYear{2025}

\begin{document}

\title{GO-N3RDet: Geometry Optimized NeRF-enhanced 3D Object Detector}
\vspace{-5mm}

\author{
Zechuan Li$^{1,2}$, Hongshan Yu$^{1}$\thanks{*Corresponding author}, Yihao Ding$^{2}$, 
Jinhao Qiao$^1$, Basim Azam$^2$, Naveed Akhtar$^2$ \\
$^1$Hunan University, $^2$The University of Melbourne \\
{\tt\small \{lizechuan, yuhongshan, qiaojh\}@hnu.edu.cn} \\
{\tt\small \{yihao.ding.1, basim.azam, naveed.akhtar1\}@unimelb.edu.au}
}
\vspace{-5mm}

\twocolumn[{%
    \renewcommand\twocolumn[1][]{#1}%
    \setlength{\tabcolsep}{0.0mm} 
    \newcommand{\sz}{0.125}  
    \maketitle
    \begin{center}
        \newcommand{\teaserwidth}{\textwidth}
    \vspace{-0.5em}
        \includegraphics[width=0.8\linewidth]{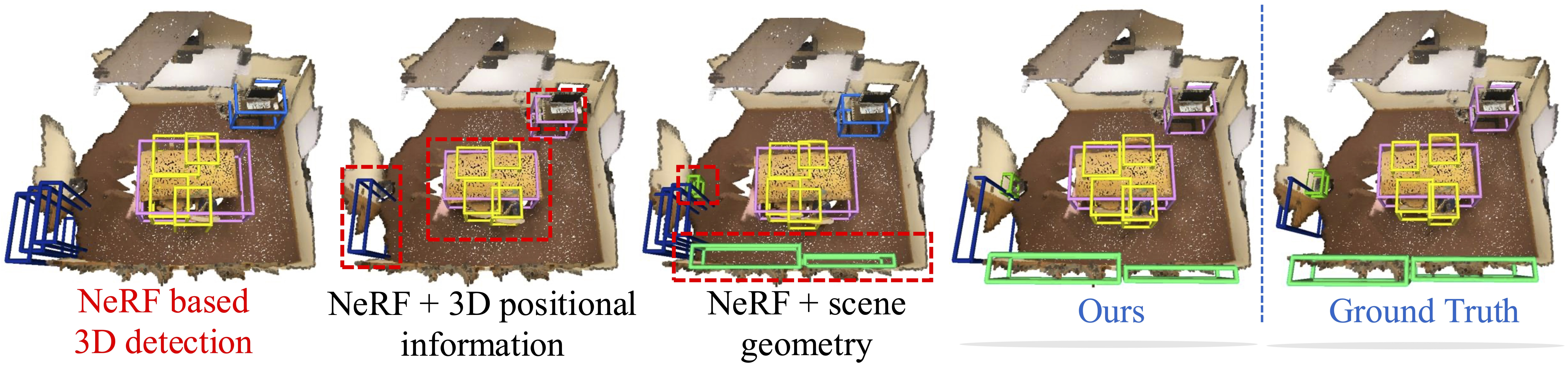}
      \vspace{-0.5em}
        \captionof{figure}{The paradigm of exploiting Neural Radiance Fields (NeRF) for multi-view 3D object detection suffers from critical issues of lack of 3D positional information modeling and insufficiency of scene geometry perception, leading to under par performance by methods like  NeRF-Det~\cite{r12} (left-most). We show that resolving these issues exclusively improves 3D object detection (dotted red boxes). 
        }
    \label{fig:fig1}
    \end{center}
}]

\maketitle
\setcounter{footnote}{0}
\renewcommand{\thefootnote}{\textsuperscript{*}}
\footnotetext{ Corresponding author: Hongshan Yu.}
\begin{abstract} 
\vspace{-2mm}
We propose GO-N3RDet, a scene-geometry optimized multi-view 3D object detector enhanced by neural radiance fields. The key to accurate 3D object detection is in effective voxel representation. 
However, due to occlusion and lack of 3D information, constructing 3D features from multi-view 2D images is challenging. Addressing that, we introduce a unique 3D positional information embedded voxel optimization mechanism to fuse multi-view features.
To prioritize neural field reconstruction in object regions, we also devise a double importance sampling scheme for the NeRF branch of our detector. We additionally propose an opacity optimization module for precise voxel opacity prediction by enforcing multi-view consistency constraints. Moreover, to further improve voxel density consistency across multiple perspectives, we incorporate ray distance as a weighting factor to minimize cumulative ray errors. Our unique modules synergetically form an end-to-end neural model that establishes new state-of-the-art in NeRF-based multi-view 3D detection, verified with extensive experiments on ScanNet and ARKITScenes. 
Code will be available at \url{https://github.com/ZechuanLi/GO-N3RDet}. 
\end{abstract}

\vspace{-5mm}
\section{Introduction}
\label{sec:intro}
\vspace{-1mm}
Multi-view 3D object detection~\cite{r11,r16,r12} uses multiple camera perspectives to enable a detailed understanding of complex indoor environments, providing richer information than monocular~\cite{r2,r3,r4} or 2D detection~\cite{r5,r41,r42,r43}. 
Employing low-cost cameras instead of costly LiDAR or depth sensors~\cite{r9,r10,r44,r45} makes  this paradigm of 3D object detection   practically appealing. Hence, its a preferred choice in indoor robot navigation~\cite{r8}, scene understanding~\cite{r51,r52}  and augmented reality ~\cite{r54,r55}. 

A critical sub-task in  multi-view 3D object detection is constructing 3D feature volumes from multi-view images. 
To that end, the pioneering approach of   ImVoxelNet~\cite{r6}  first generates a voxel grid and projects the voxel centers onto multi-view images to extract features, which are then interpolated and aggregated using average pooling across views. ImGeoNet~\cite{r11} builds further on this concept by incorporating supervisory signals to refine geometric perception from multi-view images, mitigating the ambiguity caused by free-space voxels.

\begin{figure*}[t]
    \centering
    \includegraphics[width=0.8\linewidth]{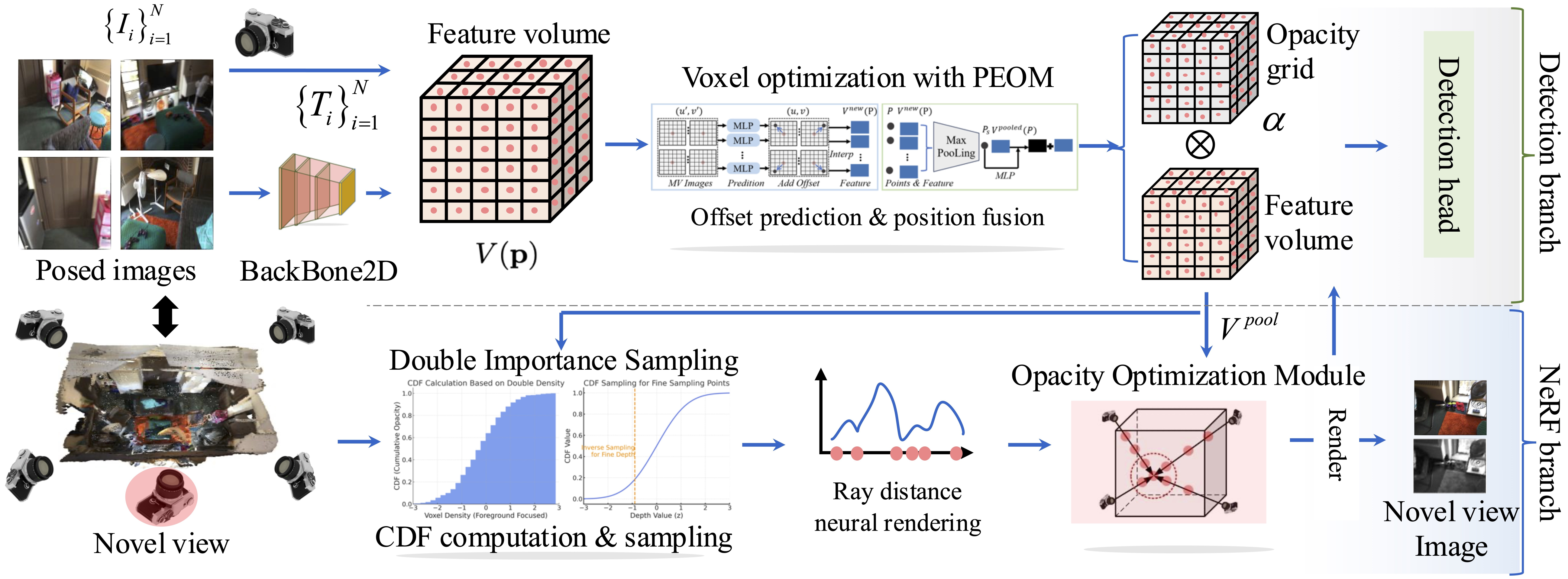}
    \vspace{-2mm}
    \caption{ \textbf{Schematics of GO-N3RDet:} Given $N$ input  multi-view images $\{I_i\}_{i=1}^N$, we first project them onto a regular 3D grid, assigning each voxel $N$ corresponding image features. The proposed Positional information
Embedded voxel Optimization Module (POEM)  fuses these features and optimizes the voxel positions. In the NeRF branch, Double Important Sampling (DIS) is employed to focus on more foreground points, and opacity is optimized using the Opacity Optimization Module (OOM). The optimized opacity is  used to adjust the fused voxel features, which are finally fed into the detection head.
    }
    \label{fig:fig2}
    \vspace{-4mm}
\end{figure*}

More recently, methods exploiting Neural Radiance Fields (NeRF)\cite{r50}
\textcolor{black}{have demonstrated significant potential to capture the geometric structure of scenes for 3D object detection} \cite{r12, r19, r20}. For instance, NeRF-Det~\cite{r12} 
predicts scene opacity through NeRF, which allows  dynamic adjustment of voxel features. By integrating the detection and NeRF branches via a shared MLP, NeRF-Det provides more accurate opacity predictions for the  voxels. 
NeRF-Det++\cite{r19} and NeRF-DetS\cite{r20} later enhanced the NeRF processing branch to further improve 3D scene understanding.
Nevertheless, the NeRF-based stream of methods  lacks in two major aspects,  mentioned  below.

(1) \textit{Deficiency of 3D positional information:} 2D image features lack the necessary spatial information for precise 3D object localization. (2) \textit{Insufficient scene geometry perception:} NeRF~\cite{r50} focuses on scene-level rendering, neglecting object-level details crucial for object detection. This results in imprecise opacity predictions, which leads to poor 3D object detection. Both problems exclusively impact the  detection performance - see Fig.~\ref{fig:fig1}.
We also verify this in our experiments \S~\ref{sec:4} (Tab.~\ref{table3}), affirming inadequate opacity estimation for the detection task.

Addressing the challenges, we propose a Geometry Optimized NeRF-enhanced 3D Object Detector (GO-N3RDet) - see Fig.~\ref{fig:fig2}.
Our technique is an end-to-end multi-view 3D object detection network that comprises multiple proposed constituent modules. 
It features a Positional information Embedded voxel Optimization Module (PEOM) to effectively construct voxels for 3D detection,  enableing  finer geometric details in voxel representation.
We also propose a Double Importance Sampling (DIS) module, which considers two density components during sampling: the density at the adjusted voxel center and the density estimated by NeRF. This enhances opacity predictions by focusing on key foreground points while ensuring broad scene coverage. Moreover, we also introduce an Opacity Optimization Module (OOM) to enforce consistency across multiple views by minimizing the variance in opacity predictions for the same location viewed from different angles, promoting smoother scene reconstruction.
In GO-N3RDet, voxel opacity is further refined by weighting multi-view opacity predictions based on ray length, reducing cumulative error and enhancing the  precision of our 3D object detector.

We  establish the effectiveness of our method by evaluating it on ScanNet~\cite{r13} and ARKitScenes~\cite{r14} datasets. Highlights of our contributions include the following.

\vspace{-2mm}
\begin{itemize}
\setlength\itemsep{-0.2em}
\item We propose GO-N3RDet -  an end-to-end trainable   network for multi-view 3D object detection.  Our model uniquely leverages positional information embedded voxels and a geometrically optimized NeRF branch.
\item We propose PEOM module to effectively embed scene positional information in 3D feature volumes, and introduce a DIS module for geometrically meaningful sampling for NeRF. Moreover, we introduce OOM module to ensure cross-perspective opacity consistency for opacity grid optimization.
\item  We establish new state-of-the-art (SOTA) in NeRF-based multi-view 3D detection  on ScanNet and ARKitScenes, \textcolor{black}{also showcasing a considerable training compute advantage over the best performing ray-matching method~\cite{r18} while retaining   performance.} 
\end{itemize}

\vspace{-3mm}
\section{Related Work}
\vspace{-1mm}

\vspace{0.5mm}
\noindent{\bf Point Cloud Based Object Detection.~}
Due to the more reliable geometric information provided by 3D point clouds, point cloud-based methods\cite{r46,r47,r48} have demonstrated promising  performance in indoor 3D object detection. These methods primarily follow two pipelines: \textit{point-based methods} and \textit{grid-based methods}. In the point-based methods, most approaches follow the pipeline introduced by VoteNet~\cite{r21}, which involves first voting for the object centers and then predicting the 3D bounding boxes and object categories. Several subsequent methods, such as MLCVNet~\cite{r22} and others~\cite{ r23, r24, r25, r9}, are improvements laregly building on the VoteNet pipeline.
Grid-based methods project  discrete and unordered 3D points onto regular voxel representations, addressing the challenges posed by the sparse and irregular nature of point clouds. For instance, 
GSDN~\cite{r28} employs sparse convolution to enhance the efficiency of 3D convolutions, with an encoder-decoder structure built from sparse 3D convolution blocks. FCAF3D~\cite{r29} builds on the basic architecture of GSDN~\cite{r28}, improving it with an anchor-free approach that claims to enhance both efficiency and performance. 
TR3D~\cite{r31} designs an efficient sparse convolutional architecture, simultaneously leveraging point cloud and image features.
While point cloud-based methods achieve strong performance, their reliance on high-cost 3D sensors significantly restricts their broader applicability.

\vspace{0.5mm}
\noindent{\bf Monocular Object Detection.~}
The field of monocular object detection has received considerable attention due to its practicality and cost-effectiveness. Numerous monocular 3D object detection methods have been developed  for outdoor autonomous driving applications~\cite{r32,r33,r34,r35,r36,r37}.
In indoor scenarios, PointFusion~\cite{r38} enhances detection accuracy by integrating depth estimation and semantic segmentation into the  detection process. COH3D~\cite{r39} and Total3D~\cite{r38} leverage three collaborative losses to jointly train 3D bounding boxes, 2D projections, and physical constraints, enabling  real-time estimation of object bounding boxes, room layouts, and camera poses from a single RGB image. Implicit3D~\cite{r40} addresses the common inaccuracies in shape and layout estimation encountered in complex and heavily occluded scenes by utilizing implicit networks. Monocular methods show promise but face challenges in accurately deriving depth and scale from a single viewpoint and are limited in handling occlusions. 

\vspace{0.5mm}
\noindent{\bf Multi-view Object Detection.~} 
ImVoxelNet~\cite{r6} was one of the first methods to construct voxel features from multiple views and apply 3D convolutional networks for object detection. However, its performance was limited by insufficient scene geometric information. Subsequent methods addressed this limitation. ImGeoNet~\cite{r11} improved scene geometry estimation by supervising voxel weights, while PARQ~\cite{r17} combined pixel-aligned appearance features with geometric information for iterative prediction refinement. NeRF-based methods, such as NeRF-RPN~\cite{r16}, use neural radiance fields to predict voxel opacity but suffer from complexity and underutilized multi-view advantages. NeRF-Det~\cite{r12} merges the NeRF branch with the detection branch for end-to-end opacity prediction, improving voxel geometry. Yet, the absence of 3D information in multi-view image features leads to inaccurate opacity predictions. Later refinements \cite{r19,r20} enhance NeRF-Det but do not directly address the fundamental problems of this type of approach. 

The recent CN-RMA~\cite{r18} method captures scene geometry using Ray Marching Aggregation to improve 3D detection. However, it relies heavily on pre-trained 3D reconstruction and ground truth truncated sign distance function for supervision, hindering end-to-end training and requiring significantly longer training times compared to NeRF-based approaches.
MVSDet~\cite{r56}  improves detection performance through plane-sweeping and soft-weighting mechanisms. However, its use of fixed depth planes in detection limits its flexibility in handling varying viewpoints, thereby weakening its implicit geometry modeling capability compared to NeRF-based methods.

\vspace{-1mm}
\section{Proposed Approach}
\vspace{-1mm}
\noindent{\bf Problem Formulation:}
We follow \cite{r6, r12} in formalizing the problem.
Let us denote the set of  $N$ images captured from different viewpoints by $\{I_i\}_{i=1}^N$. Corresponding camera parameters for each image are denoted by $\{T_i\}_{i=1}^N$, where $T_i$ includes intrinsic and extrinsic parameters of the $i$-th camera. The objective is to estimate a set of 3D bounding boxes $\{B_j\}_{j=1}^M$ for $M$ objects in the scene, where each  bounding box $B_j$ is defined by its center coordinates $(x_j, y_j, z_j)$, dimensions $(w_j, h_j, l_j)$, and orientation $\theta_j$.

For this problem, we particularly aim to overcome the limitations of the prior art in accurate perception of the scene geometry. Our GO-N3RDet approach is designed to enhance voxel quality within a scene by leveraging NeRF along with a range of proposed constituent modules for precise 3D detection. An overview of our approach is provided in Fig.~\ref{fig:fig2}, and details are given below.

\vspace{-0.5mm}
 \subsection{Feature volume construction\label{sec:3.1}}
\vspace{-1.5mm}

Our technique begins by extracting image features from each camera view using pre-trained 2D backbone network - ResNet~\cite{r53}. We express this operation as
\begin{equation}
F_{I_i} = \text{BackBone2D}(I_i),
\end{equation}
where \( F_{I_i} \in \mathbb{R}^{h \times w \times C} \) is the \( C \)-channel 2D feature map for the \( i \)-th image.
Next, we construct a voxel grid of size \( N_x \times N_y \times N_z \), where each voxel center is located at coordinate \( \mathbf{p} = (x, y, z)^\top \). We then project these voxel center coordinates onto the image feature maps of each view using the camera parameters \( T_i \), resulting in projection locations denoted as \( (u, v) \).
This projection is performed as
\begin{equation} (u, v) = \pi(T_i \mathbf{p}_h) = \left( \frac{(T_i \mathbf{p}_h)_x}{(T_i \mathbf{p}_h)_z}, \frac{(T_i \mathbf{p}_h)_y}{(T_i \mathbf{p}_h)_z} \right), 
\label{eq2}
\end{equation}
where \( \mathbf{p}_h = [x, y, z, 1]^\top \) is the homogeneous coordinate of the voxel center, \( T_i \) is the projection matrix combining both intrinsic and extrinsic parameters of the \( i \)-th camera, and \( \pi(\cdot) \) denotes the function that performs homogeneous normalization to obtain the pixel coordinates \( (u, v) \).
Once the voxel centers are projected onto the image feature maps, we perform feature interpolation at the corresponding pixel locations to obtain the voxel features as 
\begin{equation}
V(\mathbf{p}) = \text{Interpolate}\left( F_{I_i}, (u, v) \right) \in \mathbb{R}^{C}, \label{eq3}
\end{equation}
where \( V(\mathbf{p}) \) represents the voxel feature at position \( \mathbf{p} \), obtained by projecting onto the image feature map and applying interpolation. Given that there are \( N \) views and the voxel grid has dimensions \( N_x \times N_y \times N_z \), we obtain the multi-view image features corresponding to each voxel center, denoted as \( V \in \mathbb{R}^{N \times N_x \times N_y \times N_z \times C} \).

\vspace{-2mm}
\subsection{Voxel optimization with PEOM}
\vspace{-1.5mm}

Existing related methods~\cite{r6,r11,r12}  rely on average pooling to generate a representation \( V’ \in \mathbb{R}^{N_x \times N_y \times N_z \times C} \) from $V$. However, that approach has two significant limitations. {(i)} Compromised fine detail: Due to the averaging process and voxel size, the resulting voxel representation may not accurately capture the intricate details and complex geometries of object surfaces. This can lead to the loss of critical information necessary for precise 3D object detection. {(ii)} Projection errors: When projecting 3D points onto the 2D image plane, inaccuracies in camera parameters can result in projection errors, further compromising the accuracy of the representation.

\begin{figure}[t]
    \centering
    \includegraphics[width=\linewidth]{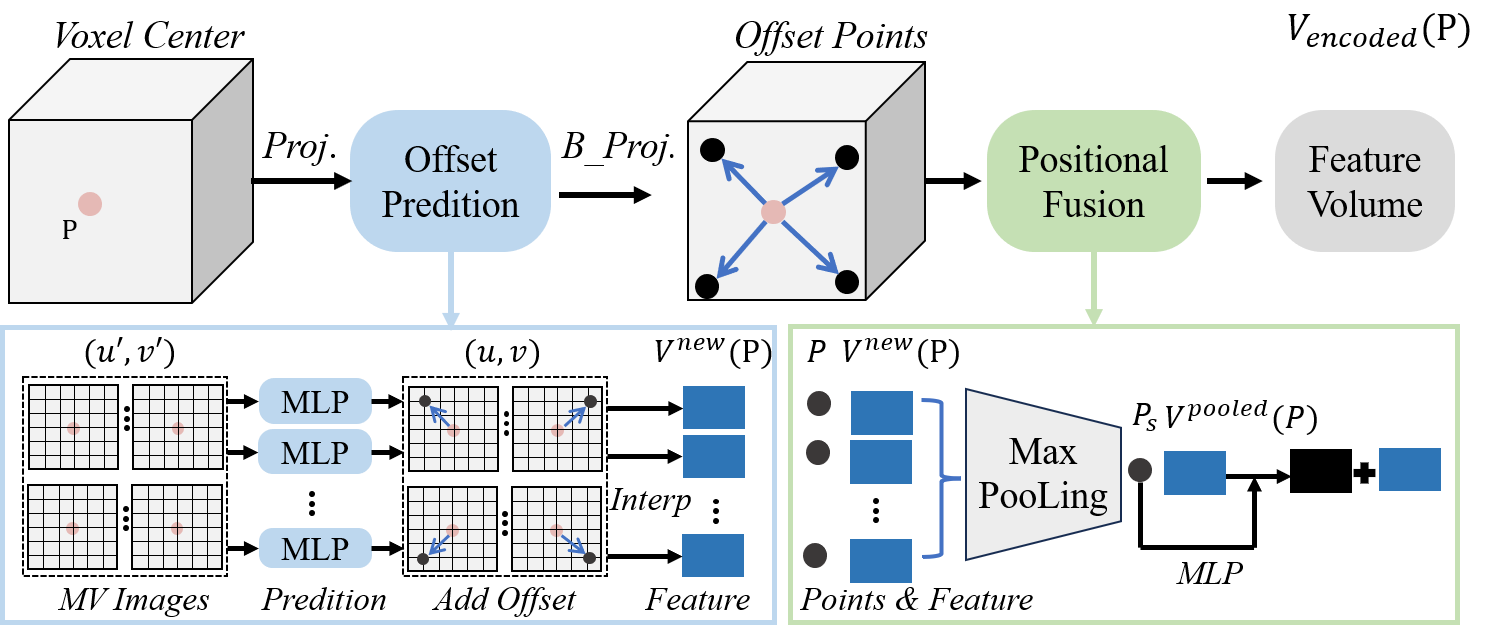}
    \caption{Illustration of \textbf{PEOM}. The voxel center is projected onto multi-view (MV) images, where each pixel coordinate predicts an offset. Features are then fused through max pooling, and the voxel position is determined based on the pixel coordinate corresponding to the maximum value. \textit{B\_Proj.} denotes back-projection.}
    \label{fig:fig3}
    \vspace{-3mm}
\end{figure} 

To address these issues, \textcolor{black}{we propose the Positional information Embedded voxel Optimization Module (PEOM), illustrated in }Fig.~\ref{fig:fig3}. This module dynamically selects projection points and optimizes voxel positions  
instead of relying solely on fixed camera parameters - details below. This enhances the method's adaptability, correcting errors caused by voxel size and camera calibration inaccuracies. The PEOM selected pixel features for the projection points more  effectively represent 3D scene information.

The proposed module projects the voxel center coordinates \( \mathbf{p} = (x, y, z)^\top \) onto the image plane using the available camera parameters \( T_i \), obtaining the corresponding pixel coordinates \( (u_i', v_i') \) based on the projection process described in Eq.~(\ref{eq2}). 
\textcolor{black}{The original voxel center can be located anywhere in space, disregarding foreground or background, and its image features may not contribute to object detection, potentially even hindering it. }Therefore, we dynamically adjust its projection on the image. For that, we predict an offset for each pixel coordinate

\begin{equation}
(\Delta u_i, \Delta v_i) = \text{MLP}\left( F_{I_i}, \mathbf{p} \right),
\end{equation}
which is  applied to the pixels  as
\begin{equation}
(u_i, v_i) = (u_i' + \Delta u_i, v_i' + \Delta v_i),
\end{equation}
Next, we obtain the corresponding pixel features \( V^\text{new} \in \mathbb{R}^{N \times N_x \times N_y \times N_z \times C} \) by interpolating the image feature maps at the adjusted pixel coordinates
\begin{equation}
V^\text{new}_i(\mathbf{p}) = \text{interpolate}\left( F_{I_i}, (u_i, v_i) \right),
\end{equation}
where \( V^\text{new}_i(\mathbf{p}) \) represents the voxel feature at position \( \mathbf{p} \) from the \( i \)-th view.
We acquire multi-view image features and perform max pooling to retain the most significant features. Max pooling fusion of features preserves the critical information within the voxel features, and compared to average pooling, it reduces noise impact, thereby enhancing the overall quality and stability of the 3D features.
\begin{equation}
V^{\text{pooled}}(\mathbf{p}) = \max_{i=1}^{N} V^\text{new}(\mathbf{p}),
\end{equation}
Here, \( V^{\text{pooled}} \in \mathbb{R}^{N_x \times N_y \times N_z \times C} \) represents the resulting feature map after max pooling.
We select the corresponding 3D coordinates by identifying the positions with the maximum responses from the encoded features:
\begin{equation}
\mathbf{p}_{\text{s}} = \arg \max_{\mathbf{p}} V^{\text{new}}(\mathbf{p}),
\end{equation}
This process ensures that the selected 3D coordinates accurately represent the scene's 3D information by utilizing multi-view features and dynamic point selection. Then, the 3D positions are encoded using an Encoder module and combined with the voxel features.
\begin{equation}
V_{\text{encoded}}(\mathbf{p}) = V^{\text{pooled}}(\mathbf{p}) + \text{Encoder}\left( \mathbf{p}_{\text{s}} \right),
\end{equation}
Finally, we obtain the optimized voxel positions \( \mathbf{p}_{\text{s}} \) and features \( V_{\text{encoded}}(\mathbf{p}) \).

\subsection{The NeRF branch\label{sec:3.3}}
\vspace{-1mm}
Constructing high-quality voxel features through multi-view information is fundamental for effective 3D object detection. However, it faces the challenge of lack of geometric information that can account for occlusions~\cite{r11}. Previous methods~\cite{r6} have relied on dense, geometry-unaware voxels, which impede accurate object detection. NeRF-Det~\cite{r12} introduces a G-MLP branch shared with a basic NeRF module to generate opacity fields that capture scene geometry. Nevertheless, their opacity estimation remains suboptimal.  Our experiments verify that the contribution of the NeRF branch to the overall detection accuracy remains limited. This happens because  (i)~such an approach  only focuses on the overall scene geometry without emphasizing the modeling of object-specific regions. To retain computational efficiency, larger voxel sizes must be chosen, and calculating the opacity of voxel center points does not accurately represent the opacity of the entire voxel region. (ii) The voxel prediction is conducted on a point-by-point basis, leading to inconsistencies across different viewpoints.
To address these issues, \textcolor{black}{we propose double important sampling and  Opacity Optimization Module (OOM)} in our NeRF branch. 

\begin{figure}[t]
    \centering
    \includegraphics[width=0.8\linewidth]{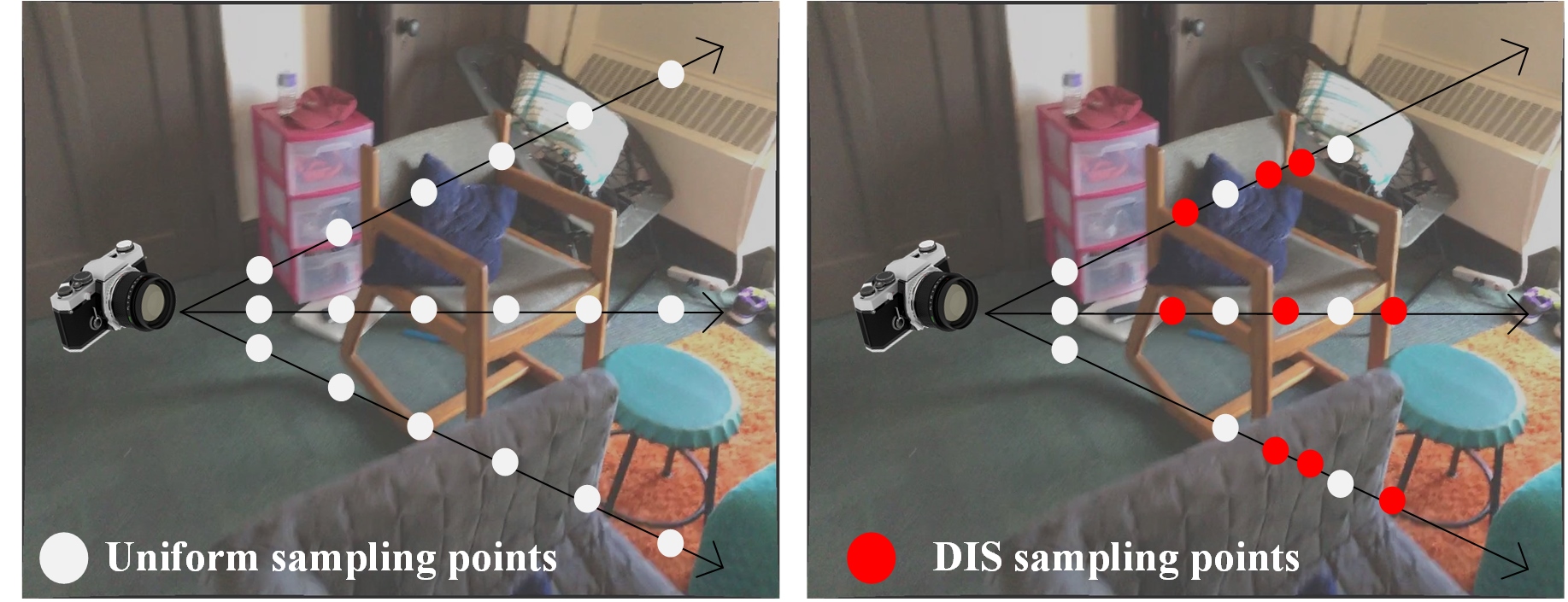}
    \caption{Illustration of \textbf{Double Important Sampling} effect. First, uniform sampling is performed, followed by CDF estimation based on the densities of the sampled points. Then, the foreground points with higher density are sampled.}
    \label{fig:fig4}
    \vspace{-5mm}
\end{figure}

\vspace{0.5mm}
\noindent\textbf{Double Important Sampling (DIS).~} \textcolor{black}{Foreground points are widely recognised for their pivotal role in point-based 3D object detection}~\cite{r9,r15}. Correspondingly, in multi-view detection, we should focus more on the foreground regions and ensure the accuracy of opacity calculation for the foreground voxels. 
To ensure that, we propose  \textcolor{black}{Double Important Sampling (DIS)} strategy, which considers both voxel density and pre-rendered density. The DIS ensures the accuracy of opacity predictions for foreground regions. 

To implement DIS, we uniformly sample along the ray, where the initial uniform sampling depth values are calculated as follows
{
\setlength{\abovedisplayskip}{3pt}
\setlength{\belowdisplayskip}{3pt}
\begin{equation}
z_i = z_{\text{near}} + i \cdot \Delta z, \quad i = 0, 1, \ldots, N_{\text{samples}} - 1,
\end{equation}
\begin{equation}
\Delta z = \frac{z_{\text{far}} - z_{\text{near}}}{N_{\text{samples}} - 1},
\end{equation}
}
This ensures that the depth sampling is evenly distributed between $z_{\text{near}}$ and $z_{\text{far}}$ with $N_{\text{samples}}$  representing the total number of samples along the ray. The sampling points are then obtained as 
\begin{equation}
\mathbf{p}_i = \mathbf{o} + z_i \mathbf{d}, 
\end{equation}
where $\mathbf{o}$ is the ray origin and $\mathbf{d}$ is the view direction. We compute the density \( \rho^m_i \) at each sampling point \( \mathbf{p}_i \) using the MLP of the NeRF branch. Additionally, based on the proximity of each sampling point to the voxel grid, we compute \( \rho^v_i \) as
{
\setlength{\abovedisplayskip}{3pt}
\setlength{\belowdisplayskip}{3pt}
\begin{equation}
\rho^v_i = \left( \frac{1}{k} \sum_{j=1}^{k} \| \mathbf{p}_i - \mathbf{p}_{i_j} \| \right)^{-1},
\end{equation}
}
where \( \mathbf{p}_{i_j} \) is the position of the \( j \)-th nearest voxel center to the sampling point \( \mathbf{p}_i \), \( k \) represents the number of nearest neighbors, and \( \| \cdot \| \) denotes the Euclidean distance.

The normalized weights of each component are first computed as follows
\textcolor{black}{\begin{equation}
\small
\hat{\rho}^m_i = \frac{\rho^m_i}{\left( \sum_{j=1}^{N_{\text{samples}}} \rho^m_j \right) + \epsilon}, \quad \hat{\rho}^v_i = \frac{\rho^v_i}{\left( \sum_{j=1}^{N_{\text{samples}}} \rho^v_j \right) + \epsilon},
\end{equation}}
where \( \epsilon \) is a small constant for numerical stability. The weight for each sampling point is computed as
{
\setlength{\abovedisplayskip}{3pt}
\setlength{\belowdisplayskip}{3pt}
\begin{equation}
w_i = \alpha \hat{\rho}^m_i + \beta \hat{\rho}^v_i,
\end{equation}
}
where \( \alpha \) and \( \beta \) are coefficients that balance the contributions of \( \hat{\rho}^m_i \) and \( \hat{\rho}^v_i \). Using the normalized weights, the Cumulative Distribution Function (CDF) is calculated as
{
\setlength{\abovedisplayskip}{3pt}
\setlength{\belowdisplayskip}{3pt}
\begin{equation}
\text{CDF}_i = \sum_{j=1}^{i} {w}_j, 
\end{equation}
}
We then generate \( N_{\text{fine}} \) uniformly distributed random variables 
$u_k \sim \mathcal{U}(0, 1):~k = 1, 2, \ldots, N_{\text{fine}}$. 
New depth values are obtained through inverse sampling of the CDF as
\begin{equation}
z^{\text{fine}}_k = \text{CDF}^{-1}(u_k), 
\end{equation}
Finally, the new 3D points are calculated as
\begin{equation}
\mathbf{p}^{\text{fine}}_k = \mathbf{o} + z^{\text{fine}}_k \mathbf{d}, 
\end{equation}
These importance-sampled points are then used for further processing. Figure~\ref{fig:fig4} illustrates the effect of DIS.

\begin{figure}[t]
    \centering
    \includegraphics[width=0.6\linewidth]{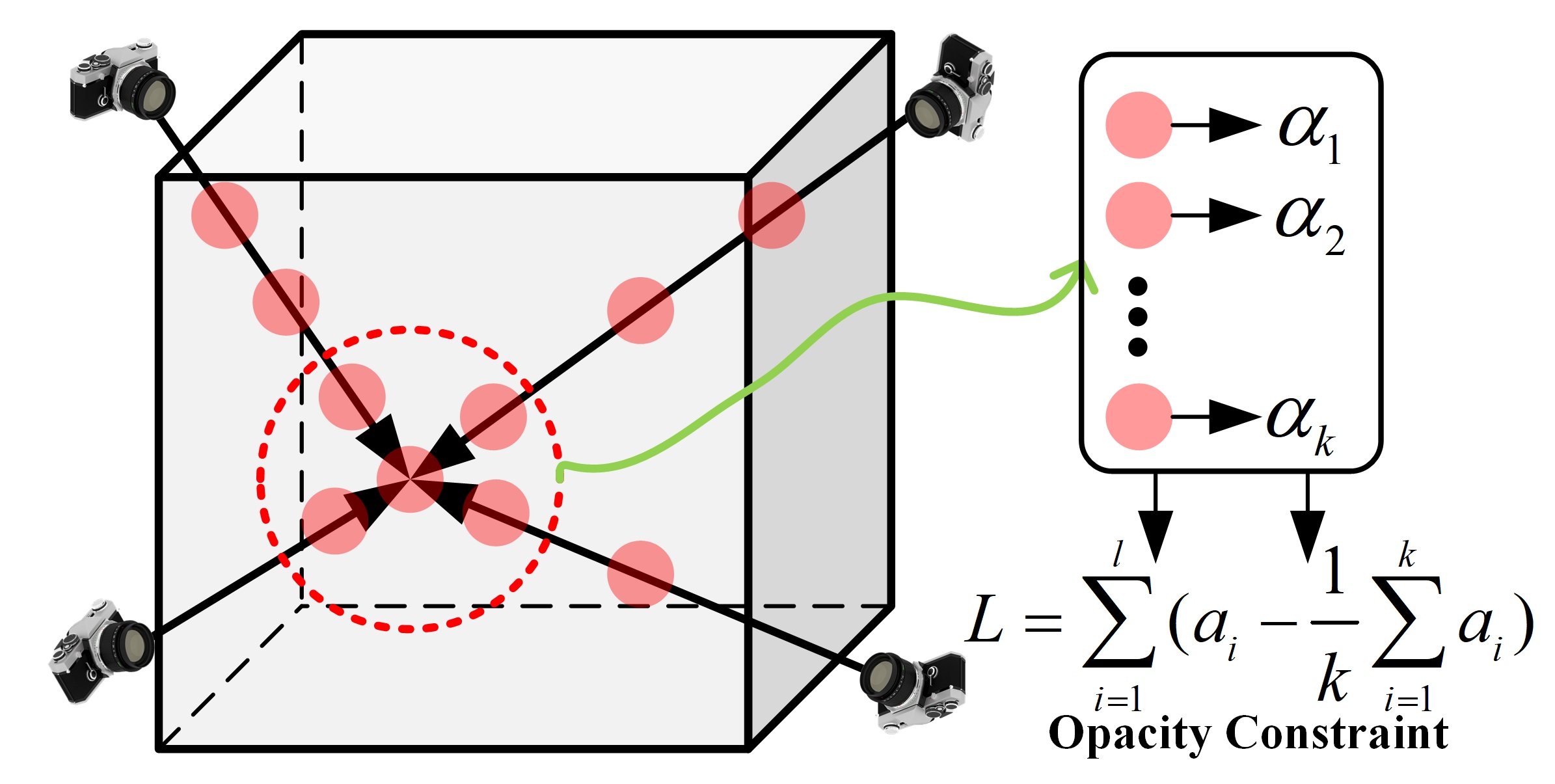}
    \caption{ Illustration of loss for \textbf{Opacity Optimization Module.}}
    \label{fig:fig5}
    \vspace{-5mm}
\end{figure}

\begin{table*}[]
\centering
{\footnotesize
\renewcommand\arraystretch{1.2}
\setlength{\tabcolsep}{0.8mm}
\begin{tabular}{c|c|cccccccccccccccccc}
\hline
\cellcolor{gray!15} ScanNet V2       & \cellcolor{gray!15}mAP@0.25  &\cellcolor{gray!15}cab  &\cellcolor{gray!15}bed  &\cellcolor{gray!15}chair  &\cellcolor{gray!15}sofa  &\cellcolor{gray!15}table  &\cellcolor{gray!15}door  &\cellcolor{gray!15}wind  &\cellcolor{gray!15} bkshf  &\cellcolor{gray!15}pic  &\cellcolor{gray!15}cntr  &\cellcolor{gray!15}desk  &\cellcolor{gray!15}curt  &\cellcolor{gray!15}fridg  &\cellcolor{gray!15}showr  &\cellcolor{gray!15}toil  &\cellcolor{gray!15}sink  &\cellcolor{gray!15}bath  &\cellcolor{gray!15}ofurn  \\ \hline
ImVoxelNet~\cite{r6}     &48.4    & 34.5 & 83.6 &72.6 &71.6 &54.2 &30.3 &14.8 &42.6 &0.8 &40.8 &65.3 &18.3 &52.2 &40.9 &90.4 & 53.3 &74.9 &33.1   \\
NeRF-Det-R50~\cite{r12}     &52.0      &37.2 &84.8 &75.0 &75.6& 51.4& 31.8& 20.0 &40.3 &0.1 &51.4 &69.1 &29.2 &58.1 &61.4 &91.5& 47.8 &75.1 &33.6  \\
NeRF-Det-R50*~\cite{r12}      &51.8      &37.7 &84.1  &74.5  &71.8  &54.2  &34.2  &17.4  &51.6  &0.1  &54.2   &71.3, &16.7  &54.5  &55.0  &92.1  &50.7  &73.8  &34.1  \\
NeRF-Det-R101*~\cite{r12}      &53.3      & 37.6 &84.9 &76.2 &76.7 &57.5 &36.4 &17.8 &47.0 &2.5 &49.2 &52.0 &29.2 &68.2 &49.3 &\cellcolor{green!15}97.1 &57.6 &\cellcolor{green!15}83.6 &35.9  \\
ImGeoNet~\cite{r11}      &54.8       &38.7 &86.5 &76.6 &75.7 &59.3 &\cellcolor{green!15}42.0 &28.1 &59.2 &4.3 &42.8 &71.5 &36.9 &51.8 &44.1 &95.2 &58.0 &79.6 &36.8  \\
NeRF-Det++$\ast$ \cite{r19}   &53.3     &38.7 &85.0 &73.2 &78.1 &56.3 &35.1 &22.6 &45.5 &1.9 &50.7 &72.6 &26.5 &59.4 &55.0 &93.1 &49.7 &81.6  &34.1  \\
NeRF-Dets$\ast$~\cite{r20}      &57.5          &43.7 &83.8 &78.3 &\cellcolor{green!15}83.0 &56.8 &43.2 &28.8 &52.0 &3.8 &70.5 &69.8 &28.5 &59.4 &61.6 &93.1 &52.4 &83.3 &44.5 
 \\
 MVSDet$\ast$~\cite{r56}        &56.2          & 40.5 & 82.4 & 79.2  & 80.2 & 55.6  & 40.3 & 25.4 & 60.9  & 3.5 & 47.3 &\cellcolor{green!15} 73.4 & 28.9 & 64.6  &\cellcolor{green!15}64.1  & 94.8 & 52.1 & 76.7 & 41.8 \\
\hline

GO-N3RDet-R50       &56.3
&40.8  &86.2  &79.9 &80.3  &55.6
&36.7  &30.4 &60.3  &4.2  &52.5 &70.4  &41.5 &53.0  &53.3 &94.8  &55.5  &78.2  &39.1  \\ 
GO-N3RDet-R50*        & 57.4  &42.3  &83.2  &78.4  &81.1  &\cellcolor{green!15}60.2 &38.8  &25.3  &56.8  &\cellcolor{green!15}4.7  &57.2 &71.2  &38.0  &\cellcolor{green!15}66.3  &59.8 &94.8  &53.5  &79.2  &40.5  \\
GO-N3RDet-R101*        &\cellcolor{green!15}58.6 &\cellcolor{green!15}44.7  &\cellcolor{green!15}87.6 &\cellcolor{green!15}80.8  &81.5  &58.2 &38.8  &\cellcolor{green!15}28.8  &\cellcolor{green!15}63.4  & 4.0  &\cellcolor{green!15} 55.4 &72.0  &\cellcolor{green!15}45.2  &58.9  &59.0 &94.8  &\cellcolor{green!15}56.5  &79.5  &\cellcolor{green!15}45.3  \\\hline
\end{tabular}
}
\vspace{-1mm}
\caption{Results on ScanNet validation set~\cite{r13} with mAP@0.25. $\ast$ denotes that the model with depth rendering supervision. R50 and R101 refer to the ResNet50 and ResNet101 backbone networks, respectively.
}\label{table1}
\vspace{-5mm}
\end{table*}

\vspace{0.5mm}
\noindent\textbf{Opacity Optimization Module (OOM).~}  When the same voxel is observed from multiple perspectives, it should maintain consistent opacity. This property is critical for accurate 3D reconstruction. 

By enforcing opacity consistency, noise introduced by viewpoint variations can be reduced, resulting in smoother and more natural 3D voxel features.

As illustrated in Fig.~\ref{fig:fig5}, a point in space is observed from $K$ different viewpoints, resulting in $K$ predicted opacities. To ensure multi-view consistency, these  opacities  must be kept consistent. To that end, we propose an opacity consistency loss as follows
\begin{equation}
\mathcal{L}_{\text{opacity}} = \frac{1}{M} \sum_{j=1}^M \sum_{i=1}^K \left( \alpha_{i,j} - \bar{\alpha}_j \right)^2,
\label{eq:opacity_loss}
\end{equation}
where $\alpha_{i,j}$ is the predicted opacity of the $j$-th point from the $i$-th viewpoint, $\bar{\alpha}_j$ is the mean opacity of the $j$-th point over all $K$ viewpoints, and $M$ is the total number of points.

In NeRF, opacity is computed by accumulating the densities of multiple samples along the direction of a light ray. As the distance along the ray increases, the number of samples traversed also increases, and small errors in each sample can accumulate, resulting in larger overall errors in opacity calculation. During propagation, light rays are subject to attenuation and scattering effects, which become more pronounced over longer distances. This implies that observations of distant points are relatively less reliable, and their opacity estimates are likely to have larger errors.

To mitigate these cumulative errors, we propose using distance as a weight to compute a weighted average of the densities. Specifically, by reducing the influence of distant points in the weighted average, the impact of cumulative errors on the final result can be minimized, thereby enhancing the reliability of the reconstruction. We compute the weighted average density (opacity) for the $j$-th point  as
\begin{equation}
\bar{\rho}_j = \frac{\sum_{i=1}^{K} w_{i,j} \rho_{i,j}}{\sum_{i=1}^{K} w_{i,j}},
\end{equation}
where the weight $w_{i,j}$ for the $i$-th viewpoint and $j$-th point is inversely proportional to the distance
\begin{equation}
w_{i,j} = \frac{1}{d_{i,j} + \epsilon},
\end{equation}
and $d_{i,j}$ is the distance from the $i$-th viewpoint to the $j$-th point, $\epsilon$ allows numerical stability, and $\rho_{i,j}$ is the density (opacity) of the $j$-th point as observed from the $i$-th viewpoint.
We adjust the voxel feature using  optimized opacity
\begin{equation}
V_{\text{adjusted}}(\mathbf{p}_j) = V_{\text{encoded}}(\mathbf{p}_j) \cdot \bar{\rho}_j,
\end{equation}
where $V_{\text{adjusted}}(\mathbf{p}_j)$ is the adjusted voxel feature at position $\mathbf{p}_j$, $V_{\text{encoded}}(\mathbf{p}_j)$ is the original encoded voxel feature at  $\mathbf{p}_j$, and $\bar{\rho}_j$ is the voxel's weighted average density (opacity).

\vspace{-1mm}
\subsection{Detection Head and Training Objective \label{sec:3.4}}
\vspace{-1mm}
To allow a fair evaluation of our contributions, we adopt the indoor detection head and related parameter settings as NeRF-Det~\cite{r12} and ImVoxelNet~\cite{r6}. It is noteworthy that we do not use the centerness loss, as we found it less effective.
We train our entire network in an end-to-end manner using a newly proposed composite loss function, defined as 
\textcolor{black}{\begin{equation}
L = L_{cls} + L_{opacity} + L_{loc} + L_{c} + L_{d}.
\end{equation}}
Here, \textcolor{black}{\(L_{cls}\) and \(L_{loc}\) represent the losses for object class prediction and bounding box prediction, respectively.} \(L_{c}\) represents the photo-metric loss. If depth maps are utilized, \(L_{d}\) represents the depth map rendering loss. \(L_{opacity}\) denotes the opacity consistency loss, with details provided in Eq.~(\ref{eq:opacity_loss}).  
 
\begin{figure*}[t]
    \centering
    \setlength{\abovecaptionskip}{0.cm}    \includegraphics[width=0.80\linewidth]{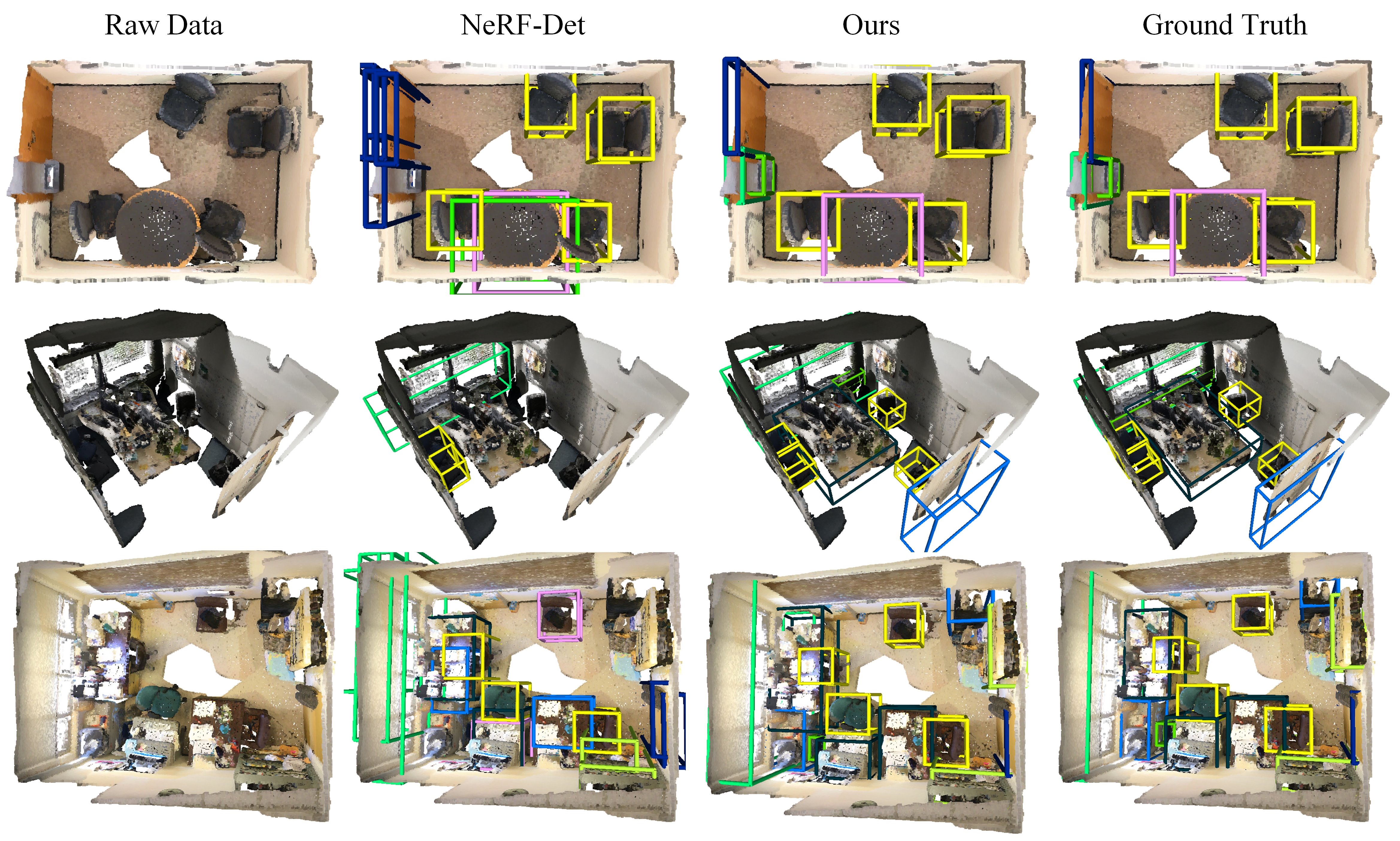}    \caption{Representative qualitative results on ScanNet  dataset~\cite{r13}. As compared to the baseline, NeRF-Det~\cite{r36}, our GO-N3RDet not only enables precise detection of more challenging objects, but also reduces false positive detections. Best viewed on screen.}
    \label{fig:fig6}
    \vspace{-3mm}
\end{figure*}  

\begin{figure}[htp]
    \centering
    \includegraphics[width=0.7\linewidth]{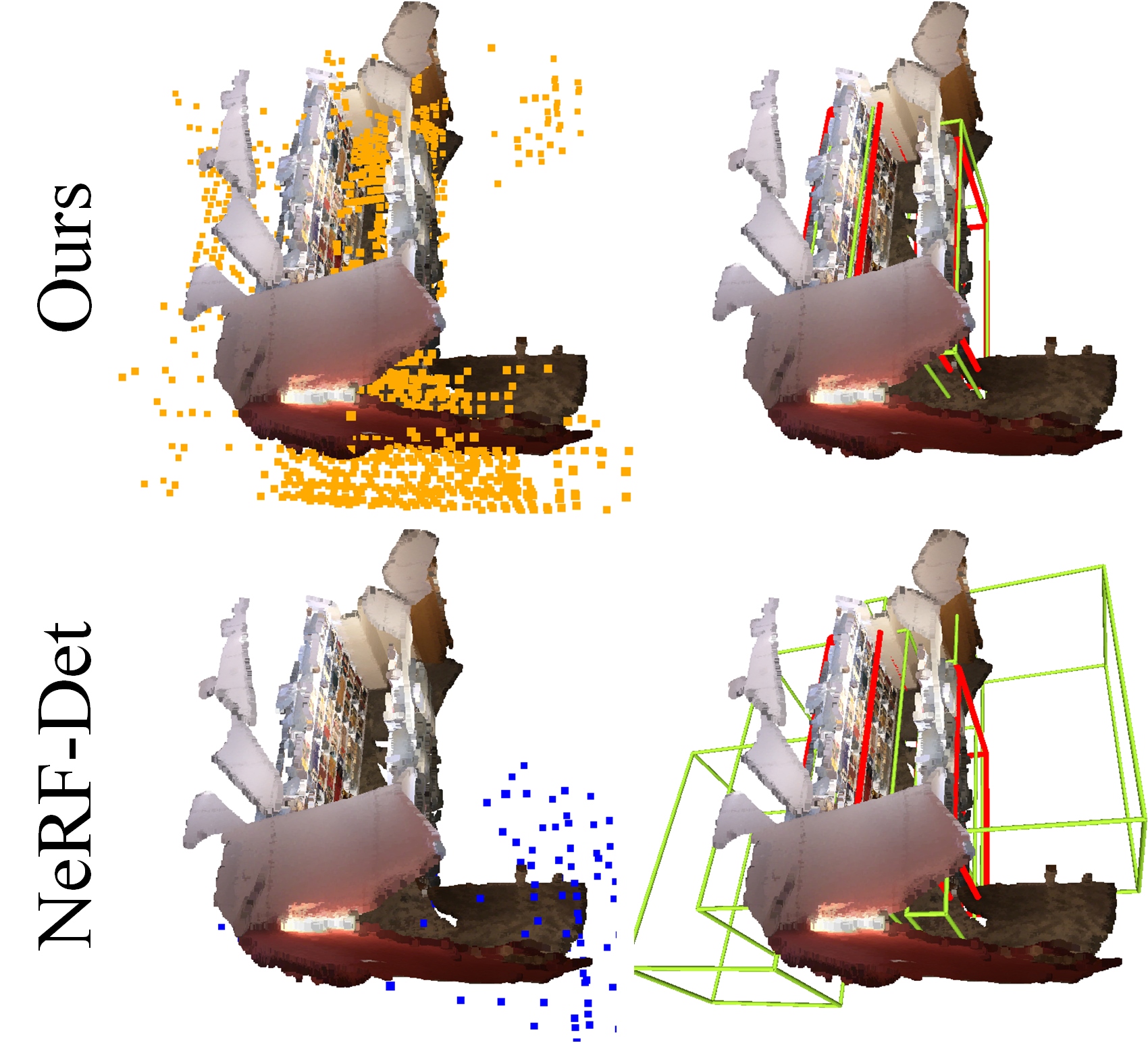}
    \caption{
    Impact of scene geometry awareness. Dots in left column indicate detected free-space. Our method predicts the free-space  accurately, leading to less false positives and precise bounding boxes in the right column.}
    \label{fig:fig7}
\vspace{-1mm}
\end{figure}

\vspace{-1mm}
\section{Evaluation\label{sec:4}}
\vspace{-1mm}
\noindent \textbf{Datasets.} 
We follow the settings of prior methods in 3D object detection and evaluate our GO-N3rDet on two indoor object detection datasets; namely, ScanNet~\cite{r13}  and ARKitScenes~\cite{r14}. ScanNet contains 1,513 complex indoor scenes with approximately 2.5M RGB-D frames. ARKITScenes~\cite{r14} contains around 1.6K rooms with more than 5,000 scans. We primarily evaluate  methods on these  datasets using mAP at 0.25 IoU and 0.5 IoU threshold.

\vspace{0.5mm}
\noindent \textbf{Implementation.} 
\textcolor{black}{Our approach integrates detection and NeRF-based scene representation within a unified framework. It is implemented using the MMDetection3D framework~\cite{r49}, optimized for 3D object detection and scene geometry perception.} 
Further implementation details are provided in the supplementary material.

\vspace{-1mm}
\begin{table}
\centering
{\small
\begin{tabular}{c|cc}
\hline
\cellcolor{gray!15}ARKITScenes     & \cellcolor{gray!15}mAP@0.25 & \cellcolor{gray!15}mAP@0.5 \\ \hline
ImVoxelNet~\cite{r6}       & 23.6     & -    \\ 
NeRF-Det~\cite{r12}   & 26.7     & 14.4   \\

\hline
 
GO-N3RDet-R50 & \cellcolor{green!15}44.7     & \cellcolor{green!15}21.9  
\\\hline
\end{tabular}
}
\caption{Results on ARKITScenes~\cite{r14} validation set.   
} 
\label{table2}
\vspace{-4mm}
\end{table}

\begin{table}[h]
\small
\centering
\begin{tabular}{c|ccc}
\hline
\cellcolor{gray!15}Method &\cellcolor{gray!15} Train epochs & \cellcolor{gray!15}Grid size & \cellcolor{gray!15}mAP@0.25 \\ 
\hline

CN-RMA~\cite{r18} & 80+12+100 & [192,192,80] & 58.6 \\ 
Ours   & 14        & [40,40,16]   & 58.6 \\ 
\hline
\end{tabular}
\caption{Comparison of CN-RMA~\cite{r18} and our method based on training epochs, grid size, and mAP@0.25.}
\label{RMA}
\vspace{-5mm}
\end{table}

\subsection{Main results\label{sec:4.2}}
\vspace{-1.5mm}

\noindent \textbf{Quantitative comparison.} As shown in Table~\ref{table1}, without relying on additional techniques, our method surpasses all NeRF-based multi-view 3D detectors on the ScanNet dataset~\cite{r13}. As compared to the foundational  multi-view method, imVoxelNet~\cite{r6}, our approach achieves an absolute 10.2\% improvement in the mAP@0.25 metric. Moreover, we demonstrate substantial improvements over three recent NeRF-based methods: NeRF-Det~\cite{r12}, NeRF-Det++\cite{r19}, and NeRF-Dets\cite{r20}.

On the ARKITScenes dataset~\cite{r14}, we again outperform existing NeRF-based methods. As illustrated in Table~\ref{table2}, our approach achieves a considerable performance gain over NeRF-Det. Given the unavailability of public results for competing methods on this dataset, we locally reproduced NeRF-Det results.
CN-RMA~\cite{r18} is not a NeRF-based method, but it has shown strong performance in multi-view tasks by utilizing Ray Marching Aggregation to capture detailed scene geometry. Given its effectiveness in these scenarios, we compared our method with CN-RMA to evaluate performance and efficiency. As the Table~\ref{RMA} shows, our approach achieves the same mAP@0.25 of 58.6 as CN-RMA, but with $\sim14$x fewer training epochs. Additionally, our method uses a much smaller grid size, demonstrating that we achieve comparable accuracy with far less computational complexity and training time. \textcolor{black}{Unlike CN-RMA, which requires ground truth TSDF~\cite{r18} supervision, our method achieves comparable performance without it, reducing reliance on costly ground truth geometry.}

\noindent \textbf{Qualitative comparison.\label{sec:4.3}} 
We compare the predicted 3D bounding boxes generated by GO-N3RDet against NeRF-Det~\cite{r12} to demonstrate the superior geometric scene perception of our approach. As shown in Fig.~\ref{fig:fig6}, GO-N3RDet consistently performs better in both simple and complex environments. 
\textcolor{black}{In the complex environment of the second row, NeRF-Det~\cite{r12} missed several key objects, including the desk, cabinet, garbage bin, and door. In contrast, GO-N3RDet successfully detected all relevant objects.} In the highly challenging scene shown in the third row, NeRF-Det~\cite{r12} exhibits multiple false positives and missed detections, while GO-N3RDet closely matches the ground truth.
Its predicted bounding boxes are  compact and accurate.
We refer to the supplementary material for more examples.

\vspace{0.5mm}
\noindent \textbf{Impact of scene geometry perception.} 
As shown in Fig.~\ref{fig:fig7}, we compare the scene geometry perception capabilities of NeRF-Det\cite{r12} and our GO-N3RDet. In the left column, the points represent free space (low-opacity areas) predicted by each method. The right column presents the ground truth (red boxes) alongside the predicted bounding boxes results (green) from both methods.
NeRF-Det~\cite{r12} detects   free space inaccurately, leading to false positives. 
In contrast, our method provides a more accurate understanding of the scene's geometry, predicting better opacity levels and, consequently, delivering more precise object detection.

\vspace{-1mm}
\subsection{Ablation Study and Discussion}
\vspace{-0.5mm}
\noindent \textbf{Effect of  sub-modules.}
We conduct an extensive ablation study to analyze the contribution of different sub-modules to our method. 
 Table~\ref{table3} reports mAP@0.25 results on the ScanNet dataset~\cite{r13}. The Naive approach refers to constructing voxel features using the method described in \S~\ref{sec:3.1}, followed by directly employing a voxel-based detection head for the detection experiments. It can be observed that when PEOM (\S~\ref{sec:3.1}), DIS (\S~\ref{sec:3.3}), and OOM (\S~\ref{sec:3.3}) are applied, each contributes to a notable improvement in the detection performance, demonstrating the effectiveness of each module. Best  results are achieved by combining them.  

\vspace{0.5mm}
\noindent \textbf{The effect of opacity.}
We also validate the effect of opacity adjustment across different methods. Opacity adjustment is crucial for perceiving scene geometry.
Table~\ref{table4} provides a quantitative analysis of the effect of opacity. \textcolor{black}{We conducted detection experiments with NeRF-Det~\cite{r12} and our GO-N3rRDet, both with and without opacity-adjusted voxel features. It can be observed that in NeRF-Det, the presence or absence of opacity adjustment does not significantly affect the final results. In NeRF-Det, we removed the opacity prediction branch, and despite not utilizing opacity to adjust voxel features, the precision at mAP@0.25 only decreased by 0.2\%. This suggests that opacity did not significantly enhance the model's ability to perceive geometric information. In contrast, in our GO-N3RDet, opacity modulation led to  3.4\% and 4.3\% improvement in precision.}

\noindent \textbf{Number of sampling points.}  We also  conducted a comprehensive study to evaluate the effect of different sampling strategies on detection performance. As shown in Tab.~\ref{table5}, increasing the number of sampling points with Uniform Sampling (Us), as employed in NeRF-Det~\cite{r12}, resulted in minimal improvements. However, even when keeping the number of sampling points fixed at 64 per ray while employing DIS, our method achieves a score of 54.2, surpassing the 53.2 achieved by uniform sampling. Furthermore, with DIS, using 128 sampling points leads to a notable performance boost, highlighting the superior efficacy of our DIS module in enhancing detection precision.

\begin{table}
\centering
{\small
\begin{tabular}{c|c|c|c|c|c}
\hline
\cellcolor{gray!15}Naive   & $\checkmark$    & $\checkmark$    & $\checkmark$  & $\checkmark$   & $\checkmark$     \\
\cellcolor{gray!15}POEM    &      & $\checkmark$     &   &  & $\checkmark$      \\
\cellcolor{gray!15}DIS       &      &      & $\checkmark$     &  &$\checkmark$     \\
\cellcolor{gray!15}OOM      &      &      &    & $\checkmark$ & $\checkmark$     \\ \hline
\cellcolor{gray!15}mAP@0.25 & 48.4 & 55.2 & 55.3 & 53.8 & \cellcolor{green!15}58.6 \\ 
\hline
\end{tabular}
}
\caption{Contribution of sub-modules (ScanNet dataset). }\label{table3}
\vspace{-3mm}
\end{table}

\begin{table}
\centering
\resizebox{\linewidth}{!}{
{\small
\begin{tabular}{c|ccccc}
\hline
\cellcolor{gray!15} & \multicolumn{2}{c}{\cellcolor{gray!15}NeRF-Det} & \multicolumn{2}{c}{\cellcolor{gray!15}Ours} \\  
\multirow{-2}{*}{\cellcolor{gray!15}Method} & \cellcolor{gray!15}mAP@0.25 & \cellcolor{gray!15}mAP@0.5 & \cellcolor{gray!15}mAP@0.25 & \cellcolor{gray!15}mAP@0.5 \\ \hline
W/O   & 53.1 & 26.9 & 55.2 & 29.4   \\
W     & 53.3(+0.2) & 27.4(+0.5) & 58.6(+3.4) &33.7(+4.3)   \\ 
\hline
\end{tabular}
}
}
\caption{Impact of opacity adjustment. ``W" refers to voxel features with opacity adjustment, while ``W/O " refers to without.}
\label{table4}
\vspace{-3mm}
\end{table}

\begin{table}[]
\centering
\small
\begin{tabular}{c|cc}
\hline
\cellcolor{gray!15}Method       & \cellcolor{gray!15}mAp@0.25
          & \cellcolor{gray!15}mAp@0.5
  \\ \hline
Us 64    & 53.2 & 27.6     \\
Us 128  &53.0 & 27.2 \\ 
DIS 64  &54.2 & 28.2 \\
 
Us 64 + DIS 64 & 54.7   & 27.8     \\
DIS 128  & 55.3  &29.2     \\\hline
\end{tabular}
\caption{Comparison of sampling methods using ScanNet dataset. ``Us'' denotes Uniform sampling. The number of samples used are mentioned in each row. DIS 128 gives the best results.}\label{table5}
\vspace{-5mm}
\end{table}

\vspace{-1mm}
\section{Conclusion }
\vspace{-1mm}
In this paper, we presented GO-N3RDet, an  end-to-end network architecture that leverages NeRF and enhance  geometric perception of multi-view voxel scenes, thereby improving the state-of-the-art for 3D object detection. The constituent PEOM module of our metho integrates multi-view features and enriches voxel representations with precise 3D positional information. To enhance opacity prediction accuracy, a DIS module focuses on foreground regions by considering the density from both NeRF predictions and voxel grid inferences, while our OOM module ensures accurate opacity calculation and consistency across multiple views. Qualitative and quantitative  results demonstrate the effectiveness of GO-N3RDet in accurately reconstructing scene geometry and improving detection reliability.

\vspace{0.75mm}
\noindent \textbf{Acknowledgement.} This work was supported by NSFC (62373140, U21A20487, U2013203), Leading Talents in Science and Technology Innovation of Hunan Province (2023RC1040), the Project of Science Fund of Hunan Province (2022JJ30024); the Project of Talent Innovation and Sharing Alliance of Quanzhou City (2021C062L). This work was completed by Zechuan Li during his visit to the University of Melbourne under the supervision of Naveed Akhtar and Hongshan yu. Zechuan Li, Hongshan Yu are with the School of Robotics, College of Electrical and Information Engineering, Quanzhou Institute of Industrial Design and Machine Intelligence Innovation, Hunan University. 

\balance
{\small
\bibliographystyle{ieee_fullname}
\bibliography{main}
}

\end{document}